\documentclass[10pt,twocolumn,letterpaper]{article}

\usepackage{wacv}              % To produce the CAMERA-READY version
\usepackage{graphicx}
\usepackage{amsmath}
\usepackage{amssymb}
\usepackage{booktabs}
\usepackage{multirow}
\usepackage{caption}
\usepackage[accsupp]{axessibility}

\usepackage[pagebackref,breaklinks,colorlinks]{hyperref}

\usepackage[capitalize]{cleveref}
\crefname{section}{Sec.}{Secs.}
\Crefname{section}{Section}{Sections}
\Crefname{table}{Table}{Tables}
\crefname{table}{Tab.}{Tabs.}

\usepackage[table, xcdraw]{xcolor}

\makeatletter
\newcommand\newtag[2]{#1\def\@currentlabel{#1}\label{#2}}
\makeatother

\newcolumntype{x}[1]{>{\centering\arraybackslash\hspace{0pt}}p{#1}}

\definecolor{codegreen}{rgb}{0,0.6,0}
\definecolor{codegray}{rgb}{0.5,0.5,0.5}
\definecolor{codepurple}{rgb}{0.58,0,0.82}
\definecolor{backcolour}{rgb}{0.968, 0.968, 0.968}

\def\wacvPaperID{2388} % *** Enter the WACV Paper ID here
\def\confName{WACV}
\def\confYear{2024}

\begin{document}

%%%%%%%%% TITLE
\title{Alleviating Foreground Sparsity for Semi-Supervised \\ Monocular 3D Object Detection}

\author{Weijia Zhang$^{1}$ \quad Dongnan Liu$^{1}$ \quad Chao Ma$^{2}$ \quad Weidong Cai$^{1}$ \\
\vspace{-10pt}\\
$^{1}$University of Sydney \quad $^{2}$Shanghai Jiao Tong University \\
{\tt\small \{wzha0649, dongnan.liu, tom.cai\}@sydney.edu.au \quad chaoma@sjtu.edu.cn}}

\maketitle

\begin{abstract}
   Monocular 3D object detection (M3OD) is a significant yet inherently challenging task in autonomous driving due to absence of explicit depth cues in a single RGB image. In this paper, we strive to boost currently underperforming monocular 3D object detectors by leveraging an abundance of unlabelled data via semi-supervised learning. Our proposed ODM3D framework entails cross-modal knowledge distillation at various levels to inject LiDAR-domain knowledge into a monocular detector during training. By identifying foreground sparsity as a main culprit behind existing methods' suboptimal training, we exploit the precise localisation information embedded in LiDAR points to enable more foreground-attentive and efficient distillation via the proposed BEV occupancy guidance mask, leading to notably improved knowledge transfer and M3OD performance. Besides, motivated by insights into why existing cross-modal GT-sampling techniques fail on our task at hand, we further design a novel cross-modal object-wise data augmentation strategy for effective RGB-LiDAR joint learning. Our method ranks 1\textsuperscript{st} in both KITTI validation and test benchmarks, significantly surpassing all existing monocular methods, supervised or semi-supervised, on both BEV and 3D detection metrics. Code will be released at \url{https://github.com/arcaninez/odm3d}.
\end{abstract}

\section{Introduction}
\label{sec:intro}
% M3OD has advantages
3D object detection represents a fundamental problem for applications in autonomous driving and robotics. Among 3D object detection from scene representations of different modalities such as LiDAR, RADAR, range images, and stereo images, monocular 3D object detection (M3OD) possesses unique advantages for practical applications. M3OD allows for easy, lightweight, and low-cost deployment on a moving platform since it only requires a single RGB camera. By performing passive sensing, cameras are also free from interference that active sensors such as LiDAR and RADAR are susceptible to, which is essential to safe autonomous driving.

\begin{figure}[t]
  \centering \captionsetup{skip=5pt}
  \begin{subfigure}{0.66\linewidth}
    \includegraphics[width=1.0\linewidth]{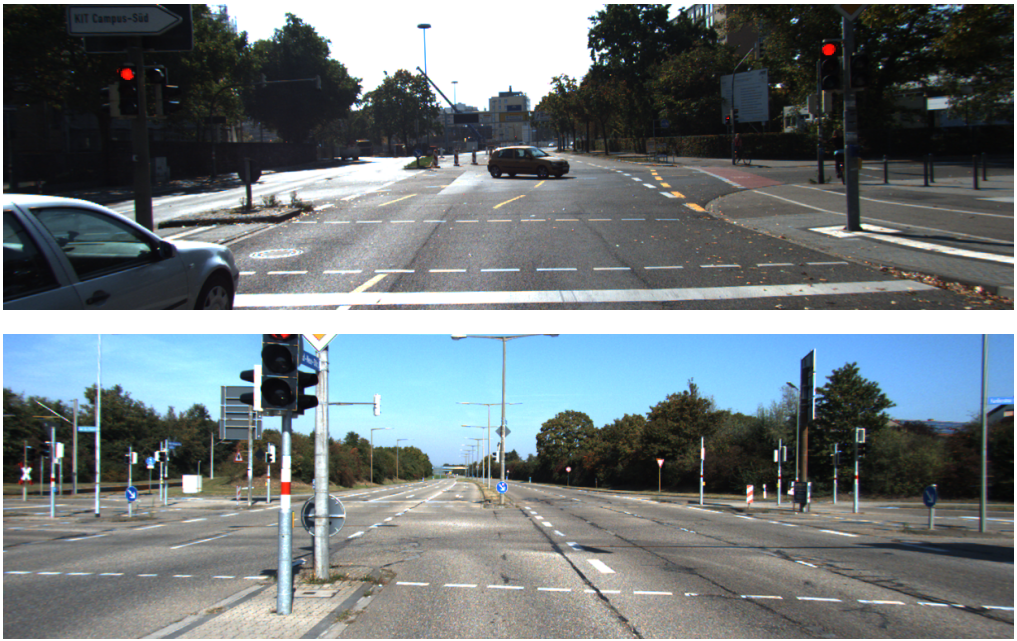}
    \caption{}
    \label{fig:inefficiency1}
  \end{subfigure}
  \hfill
  \begin{subfigure}{0.31\linewidth}
    \includegraphics[width=1.0\linewidth]{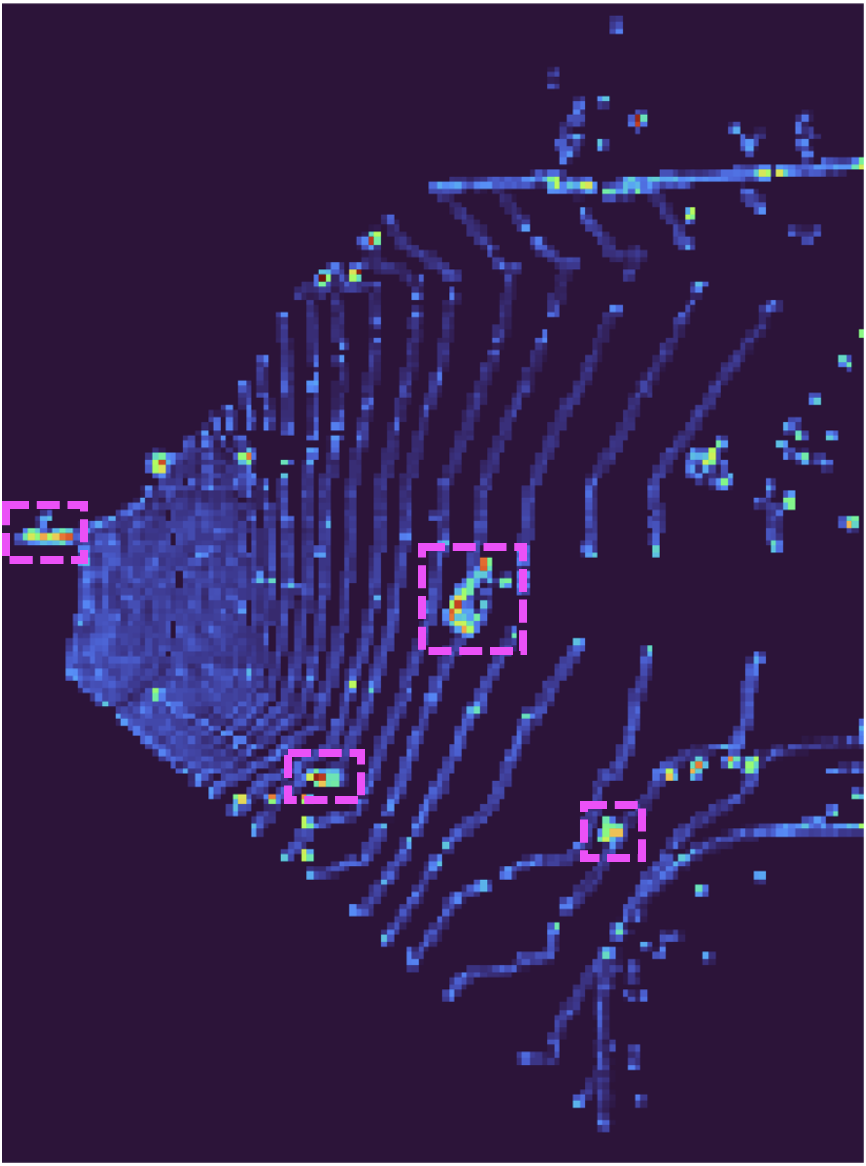}
    \caption{}
    \label{fig:inefficiency2}
  \end{subfigure}
  \caption{Two types of inefficiency identified in the state-of-the-art CMKD \cite{cmkd}: (a) object sparsity leads to insufficient training signals for the network to learn from; (b) object sparsity leads to foreground (marked with dashed boxes) signals being overwhelmed by background signals in BEV dense distillation.}
  \label{fig:inefficiency} \vspace{-6pt}
\end{figure}

% M3OD is challenging
Despite these benefits, M3OD is arguably the most challenging compared to 3D object detection receiving LiDAR, RADAR, or stereo images as input. On popular 3D object detection benchmarks such as KITTI \cite{kitti3d},  M3OD methods lag behind their LiDAR-based or stereo counterparts by a daunting margin. This is perhaps not surprising, given that an RGB image does not contain any explicit 3D measurements of a scene. Indeed, inferring 3D attributes from a single 2D image is an ill-posed problem, as pointed out in many previous works \cite{patchnet, gupnet, dd3d, simonelli21}.

% Use of depth for M3OD
To mitigate this dilemma of 2D-to-3D inference, many existing M3OD approaches resort to incorporating explicit depth estimation for enhanced depth awareness \cite{d4lcn, ddmp, caddn, dd3d}, lifting 2D images to 3D ``pseudo-LiDAR" representation via off-the-shelf depth estimators \cite{pseudolidar, m3dplidar, am3d, simonelli21}, or directly utilising matched depth maps in training for RGB-depth feature fusion \cite{monodtr, monodetr, add}. 
% Use of priors for M3OD
Other methods tackle the ill-posed M3OD problem by imposing geometric constraints such as inter-keypoint \cite{rtm3d, monojsg, monorun, monocon} or inter-object \cite{monopair, pgd} relations to regularise 3D predictions. These have led to consistent and incremental improvements on M3OD.

% Use of SemiSL for M3OD
Parallel to advancements in M3OD, semi-supervised learning (SemiSL) has emerged as a powerful paradigm that enables learning from additional unlabelled data. Motivated by its success, in this paper we advocate exploiting large amounts of unlabelled data to boost M3OD performance. Among preliminary works on semi-supervised M3OD \cite{lpcg, mixteaching, cmkd}, CMKD \cite{cmkd} employs a simple cross-modal knowledge distillation framework to acquire the capability of learning from both images and LiDAR point clouds, labelled and unlabelled, delivering state-of-the-art monocular detection performance. 

% Drawbacks in M3OD
Despite its impressive results, upon in-depth investigation we made two insightful observations as to how CMKD lacks efficiency in its training as a result of \textit{foreground sparsity} (illustrated in Fig. \ref{fig:inefficiency}):
% Drawback 1
(i) Autonomous driving scenes often contain too few or even no objects of interest, leading to insufficient training signals for the network (Fig. \ref{fig:inefficiency1}). In particular, CMKD utilises large amounts of unlabelled samples from the KITTI Raw dataset~\cite{kittiraw}, in which scenes containing no objects at all are common.
% Drawback 2
(ii) In dense distillation, object sparsity results in foreground signals being overwhelmed by the background noise of a much larger area in bird's-eye view (BEV) (Fig. \ref{fig:inefficiency2}), which undermines accurate feature extractions for cross-modal learning.

% Our solutions
To mitigate (i), we design a novel object-wise cross-modal data augmentation technique to paste additional objects into training scenes. GT-sampling-based \cite{second} cross-modal augmentation \cite{pointaugmenting, moca, vff} has recently been employed by several multi-modal 3D object detectors \cite{vff, focalsconv, logonet}. However, these strategies are limited since their augmented scenes are produced via IoU-based collision tests, which fail to consider the relative depth of objects, as discussed in Sec.~\ref{sec:ablation_cmaug}. To alleviate this issue, we propose an occlusion-aware cross-modal GT-sampling strategy to augment the training scenes for enhanced RGB-LiDAR distillation.

For (ii), inspired by foreground-attentive distillation in 2D \cite{fgfi, fgd, ld} and 3D object detection \cite{monodistill, unidistill, ligastereo}, our proposed distillation method focuses on regions where objects more likely exist rather than treating all locations indifferently. In the absence of ground-truth labels indicating where objects are, we resort to the underlying point location knowledge embedded in point clouds, which serves as an implicit indicator of where objects and foreground might be. Intuitively, locations containing LiDAR points more likely contain an object or part of an object, and vice versa. Hence, we propose to exploit LiDAR point occupancy as a guidance for distillation in BEV.

Our designs effectively alleviate aforementioned issues caused by foreground sparsity, leading to a top-performing M3OD framework based on cross-modal distillation and semi-supervised learning. Besides, our designs are only involved in training and therefore do not introduce any additional computational or memory overhead at inference.

In summary, our contributions include: \vspace{-5pt}
\begin{enumerate}
    \item We propose occupancy-guided cross-modal distillation for M3OD, utilising the underlying localisation information in LiDAR as guidance for foreground-attentive knowledge transfer. \vspace{-8pt}
    \item  We design CMAug, a new and versatile cross-modal augmentation strategy built upon a novel occlusion-aware collision criterion, which suits both supervised and semi-supervised learning settings. \vspace{-8pt}
    \item Our ODM3D framework achieves 1\textsuperscript{st} place in KITTI \textit{val} and \textit{test} benchmarks, in term of both $AP_{3D}$ and $AP_{BEV}$, among all published supervised and semi-supervised monocular methods. 
\end{enumerate}

\begin{figure*}[t!]
  \centering \captionsetup{skip=5pt}
  \includegraphics[width=0.85\linewidth]{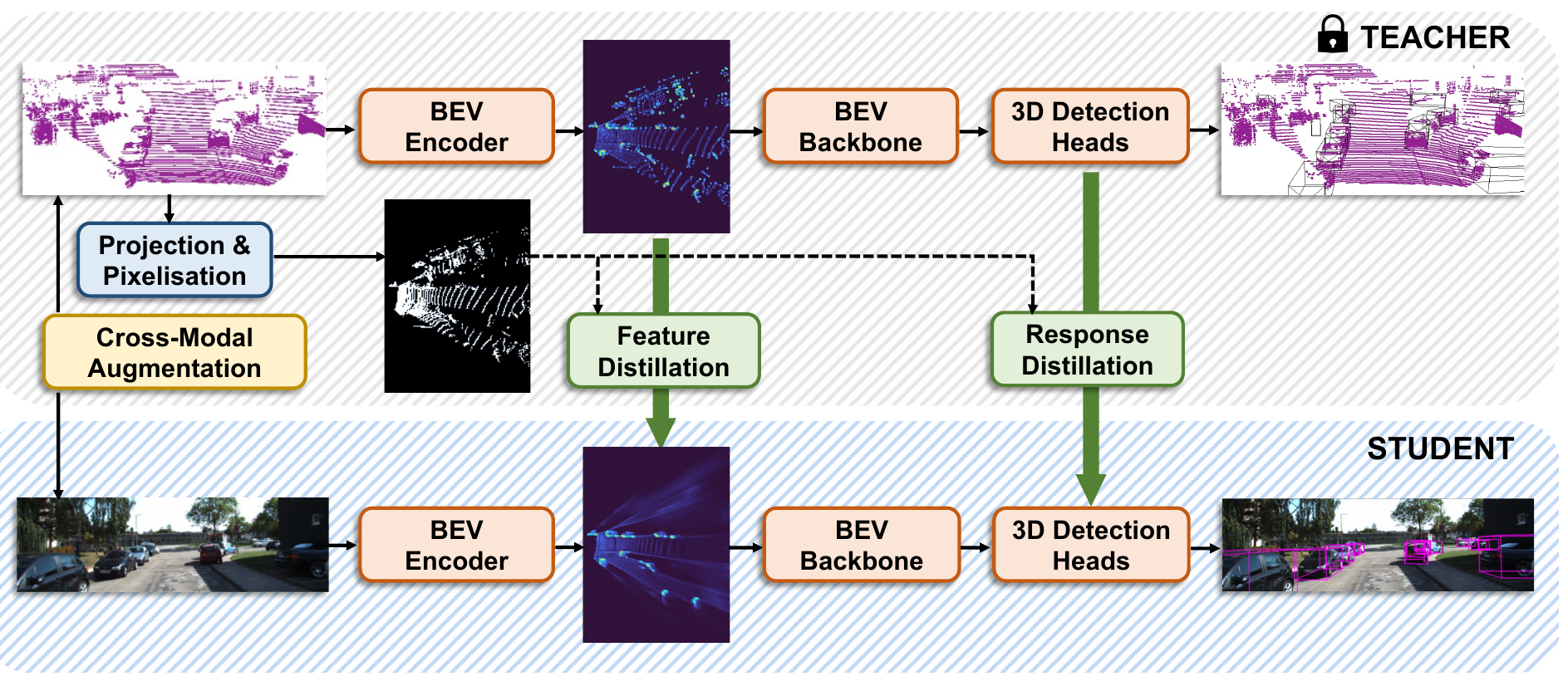}
  \caption{The ODM3D framework. Knowledge distillation is conducted in both  feature and prediction spaces in BEV, both guided by a BEV occupancy map derived from ground-truth point clouds. The teacher model is frozen during knowledge distillation, and is discarded at inference. All modules and operations within the grey shaded region are not involved at inference.}
  \label{fig:framework} \vspace{-5pt}
\end{figure*}

\section{Related Work}
\subsection{Monocular 3D Object Detection}
Monocular 3D object detection (M3OD) methods can be primarily categorised into image-only, geometric-pior-assisted, and depth-assisted methods. Image-only methods \cite{m3drpn, smoke, fcos3d, monodle, monodis} directly regress objects' 3D attributes from an RGB image. Prior-assisted methods introduce complex geometric constraints in forms of keypoint \cite{rtm3d, monojsg, monorun, monocon}, inter-object relational \cite{monopair, pgd}, camera extrinsic \cite{monoef} and temporal \cite{kinematic3d} regularisation. Depth-assisted methods make explicit use of depth to alleviate inherent depth ambiguity in images. Among them, some \cite{pseudolidar, m3dplidar, am3d, simonelli21, end2endpl} leverage off-the-shelf depth estimators (\eg DORN \cite{dorn}) to convert images into a ``Pseudo-LiDAR" representation, on which standard LiDAR-based detectors can be applied; some \cite{d4lcn, ddmp, caddn, dd3d} introduce depth estimation as an auxiliary task to learn depth-aware features for accurate 3D inference through RGB-depth fusion or multi-task learning; others \cite{mlf, pstereo} exploit depth in the form of disparity maps.
Besides, uncertainty modelling is commonly adopted \cite{monopair, monorun, gupnet, monoflex, monodle} for more accurate and robust estimation of 3D attributes. More recently, transformer \cite{transformer, detr} has been utilised for more effective contextual and depth-aware feature aggregation \cite{monodtr, monodetr, add, monoatt, mogde}. A few methods \cite{sgm3d, monodistill, add} also utilise external data during training via knowledge distillation, which are detailed in Sec.~\ref{sec:related_work_kd}. 

\subsection{Knowledge Distillation for M3OD}
\label{sec:related_work_kd}
A popular technique to transfer knowledge from a stronger model to a weaker one, knowledge distillation \cite{kd} has been under-explored in the context of M3OD. Among early efforts, SGM3D \cite{sgm3d} distills the knowledge of a teacher trained with stereo images to a monocular CaDDN \cite{caddn} student; MonoDistill \cite{monodistill} and ADD \cite{add} have their teacher and student based on an identical architecture. MonoDistill's teacher directly takes as input LiDAR-projected depth maps and guides a MonoDLE \cite{monodle} detector. In contrast, ADD's teacher receives the depth maps as extra input, and is shown to boost multiple monocular detectors \cite{patchnet, caddn, monodetr}. Contrary to these approaches, we directly employ a LiDAR-based teacher and distill stronger, more 3D-aware knowledge learnt from raw point clouds.

\subsection{Semi-Supervised M3OD}
Semi-supervised learning (SemiSL) enables learning from both labelled and unlabelled data. In the context of M3OD, KM3D \cite{km3d}, MVC-MonoDet \cite{mvcmonodet}, and Lian \etal \cite{geomconsist} enforce consistency in terms of object keypoints \cite{km3d}, bounding box predictions\cite{mvcmonodet, geomconsist}, or object-level photometry \cite{mvcmonodet} between teacher and student responses.
More akin to our method are pseudo-labelling-based approaches \cite{lpcg, mixteaching, cmkd}, which employ a teacher model pre-trained on labelled data to produce predictions (\ie pseudo-labels) for unlabelled data. Instead of directly using the teacher's detection results as in LPCG \cite{lpcg} and Mix-Teaching \cite{mixteaching}, CMKD \cite{cmkd} lets the student learn the teacher's intermediate features and dense prediction maps via knowledge distillation, achieving state-of-the-art M3OD performance. 

\subsection{Cross-Modal Data Augmentation}
Data augmentation has been a major driver behind the enormous success of deep learning. In LiDAR-based 3D object detection, GT-sampling \cite{second} pastes ground-truth object points into training scenes to diversify and proliferate objects that can be used to train the detector, and is widely adopted by subsequent LiDAR-based detectors \cite{pointpillars, pvrcnn, votr, pdv}. However, extending it to RGB-LiDAR cross-modal learning tasks is less straightforward due to difficulties in maintaining scene-level consistency between augmented RGB and LiDAR data. Recently, several cross-modal augmentation strategies based on GT-sampling have been proposed \cite{pointaugmenting, autoalignv2, vff}. They all crop and paste image regions corresponding to pasted object points, and conduct collision tests to avoid severe overlapping in perspective view (PV), with promising results yielded on recent multi-modal 3D object detectors \cite{vff, focalsconv, transfusion, logonet}. Yet, to our best knowledge, such strategies have not been explored for cross-modal distillation and semi-supervised learning. In this work, we show that existing strategies lead to augmented scenes extremely challenging if not infeasible for the monocular detector to learn from, and alleviate the issue with our proposed designs.

\section{Methodology}
\subsection{Overall Framework}
Our proposed ``Occupancy-Guided Distillation for Monocular 3D Object Detection" (ODM3D) framework follows a teacher-student paradigm with cross-modality knowledge distillation, as shown in Fig. \ref{fig:framework}. The teacher is a pre-trained LiDAR-based 3D object detector which produces intermediate BEV features within its pipeline and performs subsequent 3D object detection in the BEV space. The student is a monocular detector which takes as input a single RGB image and also involves intermediate BEV features. It is trained to mimic the teacher's intermediate BEV features at its BEV encoder and dense prediction maps at its detection heads. In this process, the student acquires LiDAR-induced knowledge from the teacher. Throughout the cross-modality training, a BEV occupancy mask obtained by projecting each scene's point cloud (detailed in Sec.~\ref{sec:occupancy_mask}) is used to guide distillation in both feature and prediction domains (Sec.~\ref{sec:okd}). Due to object sparsity in training scenes, we design and apply cross-modal data augmentation, pasting ground-truth objects into each training scene to enrich supervisory signals (Sec.~\ref{sec:cmaug}). At inference, the LiDAR-based teacher is discarded and only the monocular student is deployed.

\subsection{LiDAR-Projected BEV Occupancy Mask}
\label{sec:occupancy_mask}
Given raw point cloud $\mathbf{L}$ in a continuous 3D domain, we first project the voxelised $\mathbf{L}$ into BEV to obtain a 3D BEV map $\mathbf{M'} \in \mathbb{R}^{W_{\mathrm{BEV}} \times H_{\mathrm{BEV}} \times D}$ that has the same width $W_{\mathrm{BEV}}$ and height $H_{\mathrm{BEV}}$ as intermediate BEV features of the teacher and student networks used for feature distillation. Next, we perform ``pixelisation" of $\mathbf{M'}$. Specifically, we consider BEV representation $\mathbf{M'}$ an indicator of point occupancy status in the 3D space. Each element in $\mathbf{M'}$ can be regarded as a grid that corresponds to a 3D volume in the voxelised LiDAR space. We let an element in $\mathbf{M'}$ equal to one if its corresponding 3D volume in the LiDAR space contains at least one point, and zero if the 3D volume contains no points. Consequently, a ``one" grid represents an active occupancy grid and a ``zero" grid represents an empty occupancy grid. Afterwards, we collapse $\mathbf{M'}$ along dimension $D$ to form our 2D BEV occupancy mask $\mathbf{M}_{\mathrm{OCC}} \in \mathbb{R}^{W_{\mathrm{BEV}} \times H_{\mathrm{BEV}}}$. Concretely, an element in $\mathbf{M}_{\mathrm{OCC}}$ is one if there is at least one active grid among all $D$ grids at this location in $\mathbf{M'}$; an element in $\mathbf{M}_{\mathrm{OCC}}$ is zero if all $D$ grids at this location in $\mathbf{M'}$ are empty. 

In our experiments, voxelised point cloud $\mathbf{L}$ has a shape of $(W=1{\small,}120,\, H=1{\small,}540,\, D=40)$, which is determined by our choice of voxelisation resolution and point cloud range, and the intermediate BEV feature has $W_{\mathrm{BEV}}=140$ and $H_{\mathrm{BEV}}=188$. Hence, a grid in the proposed occupancy mask corresponds to a total of $8 \times 8 \times 40 = 2{\small,}560$ voxels, equivalent to a cubic volume of dimension $[0.32m, 0.32m, 4m]$.

\begin{figure}[t]
    \centering
    \begin{minipage}{0.478\textwidth}
        \centering
        \includegraphics[width=\linewidth]{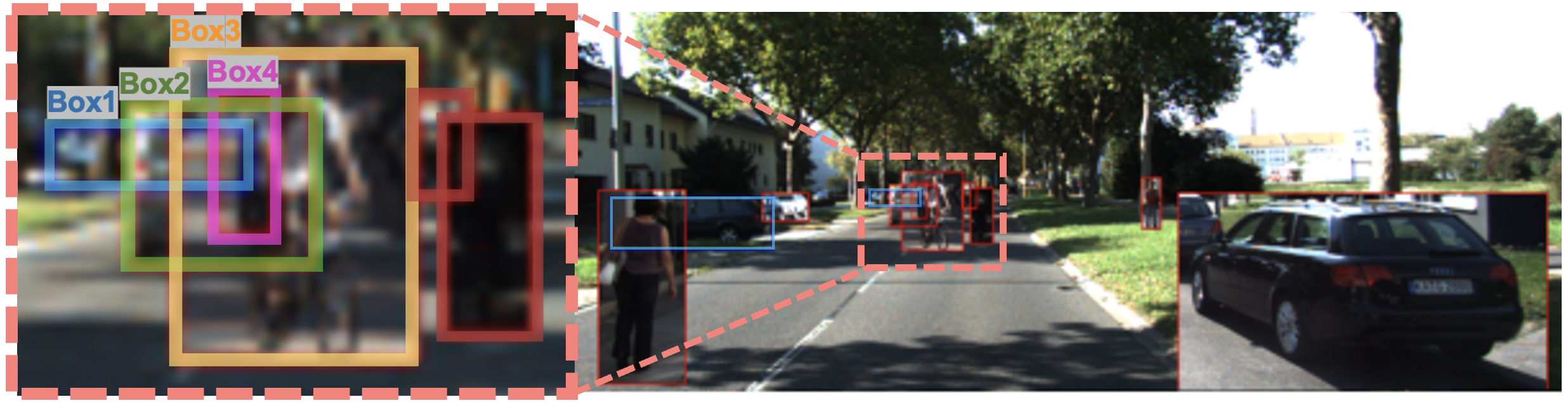}
        \label{fig:figure1} \vspace{-5pt}
    \end{minipage} 
    \begin{minipage}{0.478\textwidth}
        \centering
        \includegraphics[width=\linewidth]{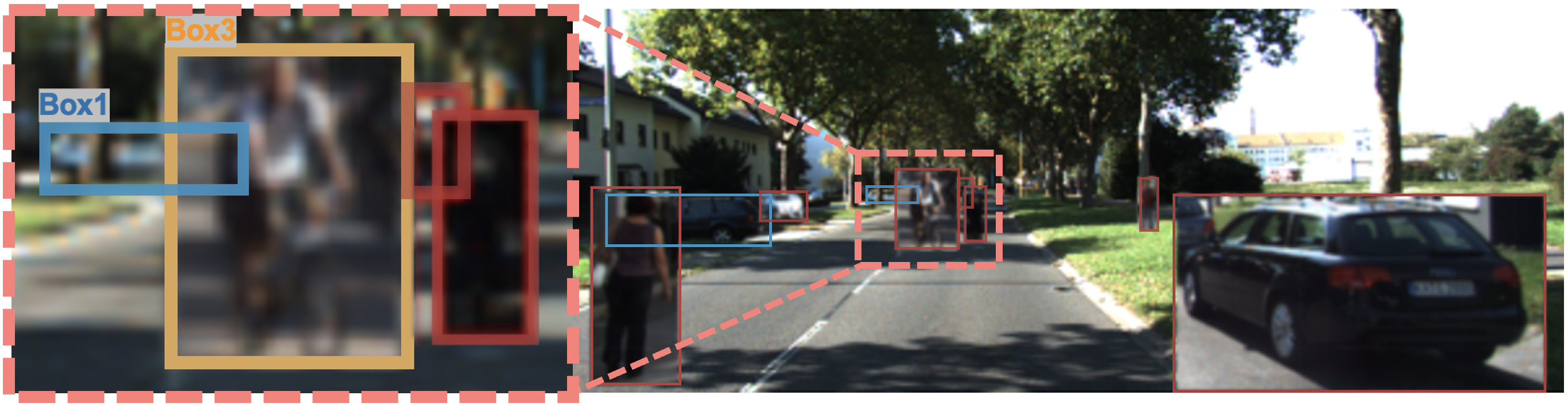}
        \label{fig:figure2}  \vspace{-15pt}
    \end{minipage}
    \caption{Examples of the same scene augmented with collision tests using IoU and our proposed OAIS thresholds.}
    \label{fig:cmaug} \vspace{-5pt}
\end{figure}

\subsection{Occupancy-Guided Knowledge Distillation} \label{sec:okd}
\noindent \textbf{Occupancy-guided feature distillation.}
Feature-level distillation is carried out between intermediate BEV features of the teacher and the student, $\mathbf{F}^{\mathrm{Tea}}_{\mathrm{BEV}} \in \mathbb{R}^{W_{\mathrm{BEV}} \times H_{\mathrm{BEV}} \times C_{\mathrm{BEV}}}$ and $\mathbf{F}^{\mathrm{Stu}}_{\mathrm{BEV}} \in \mathbb{R}^{W_{\mathrm{BEV}} \times H_{\mathrm{BEV}} \times C_{\mathrm{BEV}}}$, respectively. The mean square error (MSE) loss is adopted as feature distillation loss $\mathcal{L}_{\mathrm{Feat}}$, on which the proposed 2D BEV occupancy mask $\mathbf{M}_{\mathrm{OCC}}$ is imposed, guiding the distillation to focus on foreground regions while ignoring unimportant and interfering background. In addition, we apply Gaussian smoothing to the BEV occupancy mask which is originally binary. Converting a hard occupancy mask to a soft one effectively mitigates potential misalignment between occupancy maps and feature maps incurred in calibration, projection, or feature extraction. Mathematically, the occupancy-guided feature distillation loss is given by:
\begin{equation} \small
    \mathcal{L}_{\mathrm{FeatKD}} = \| (\mathbf{G}(\sigma) \circledast \mathbf{M}_{\mathrm{OCC}}) \odot 
    \mathcal{L}_{\mathrm{Feat}}(\mathbf{F}^{\mathrm{Stu}}_{\mathrm{BEV}}, \mathbf{F}^{\mathrm{Tea}}_{\mathrm{BEV}}) \|^{2}
    \label{eq:featkd}
\end{equation}
where $\mathbf{G}(\sigma)$ is a Gaussian kernel with standard deviation $\sigma$ determined by the kernel size; $\circledast$ is the convolution operator; $\odot$ denotes the Hadamard product operator; the process of broadcasting $\mathbf{M}_{\mathrm{OCC}}$ to match the channel dimension of $\mathbf{F}^{\mathrm{Tea}}_{\mathrm{BEV}}$ and $\mathbf{F}^{\mathrm{Stu}}_{\mathrm{BEV}}$ is omitted here for brevity. 

\noindent \textbf{Occupancy-guided response distillation.}
Inheriting the spirit of occupancy-guided feature distillation, we further impose occupancy guidance in the response space (\ie dense predictions). This design is made feasible and rational by the fact that predictions of both our teacher and student, along with pre-defined anchors, are made in the BEV space, carrying similar physical connotation to BEV features and BEV occupancy masks. As such, we again apply the BEV occupancy mask on the dense prediction maps generated by both the teacher and student networks.

Nevertheless, considering how anchors and boxes are defined and carried on these dense maps, we argue that it would be more desirable to adopt slackened occupancy guidance rather than stringent, pixel-to-pixel dictation. Essentially, the validity of direct pixel-wise multiplication between a BEV occupancy mask and a BEV feature mask stems from a pixel-to-pixel correspondence between the two representations (even though we have chosen to slacken this correspondence), which however does not hold between a BEV occupancy mask and boxes or anchors defined on prediction maps. It is perfectly normal for the centre of an anchor not to be in the immediate vicinity of points of the ground-truth object to which the anchor is matched. In light of this, we once again opt for Gaussian smoothed occupancy masks, whose benefits will be shown in Sec.~\ref{sec:ablation}.

Our teacher and student both adopt SSD-style \cite{ssd} detection heads, comprising a classification head, a localisation head, and a direction head. The occupancy-guided distillation loss for the classification head is given by:
\begin{equation} \small
    \mathcal{L}_{\mathrm{ClsKD}} = \| (\mathbf{G}(\sigma) \circledast \mathbf{M}_{\mathrm{OCC}}) \odot \mathcal{L}_{\mathrm{Cls}}(\mathbf{P}^{\mathrm{Stu}}_{\mathrm{cls}}, \mathbf{P}^{\mathrm{Tea}}_{\mathrm{cls}}) \|^{2}
\end{equation}
where $\mathbf{P}^{\mathrm{Stu}}_{\mathrm{cls}}$ and $\mathbf{P}^{\mathrm{Tea}}_{\mathrm{cls}}$ are student's and teacher's classification prediction maps, and other symbols follow Eqn. \ref{eq:featkd}.
The localisation and direction distillation losses (\ie $\mathcal{L}_{\mathrm{LocKD}}$ and $\mathcal{L}_{\mathrm{DirKD}}$), based on $\mathcal{L}_{\mathrm{Loc}}$ and $\mathcal{L}_{\mathrm{Dir}}$ respectively, are defined likewise and omitted for brevity. We adopt the quality focal loss (QFL) \cite{qfl}, smooth L1 loss, and cross-entropy loss for $\mathcal{L}_{\mathrm{Cls}}$, $\mathcal{L}_{\mathrm{Loc}}$, and $\mathcal{L}_{\mathrm{Dir}}$, respectively. The occupancy-guided response distillation loss $ \mathcal{L_{\mathrm{RespKD}}}$ is a weighted sum of distillation losses of all three detection heads. Finally, the overall distillation loss to train the teacher-student framework is a weighted sum of $\mathcal{L_{\mathrm{FeatKD}}}$ and $\mathcal{L_{\mathrm{RespKD}}}$. Note that while a concurrent work \cite{bevlgkd} also exploits LiDAR-projected masks as guidance, we stress that our occupancy guidance generalises to both feature and response domains and is used in a different setting (\ie cross-modal distillation and semi-supervised learning) and task (\ie M3OD).

\begin{figure}[t]
    \centering \captionsetup{skip=5pt} \includegraphics[width=0.95\columnwidth]{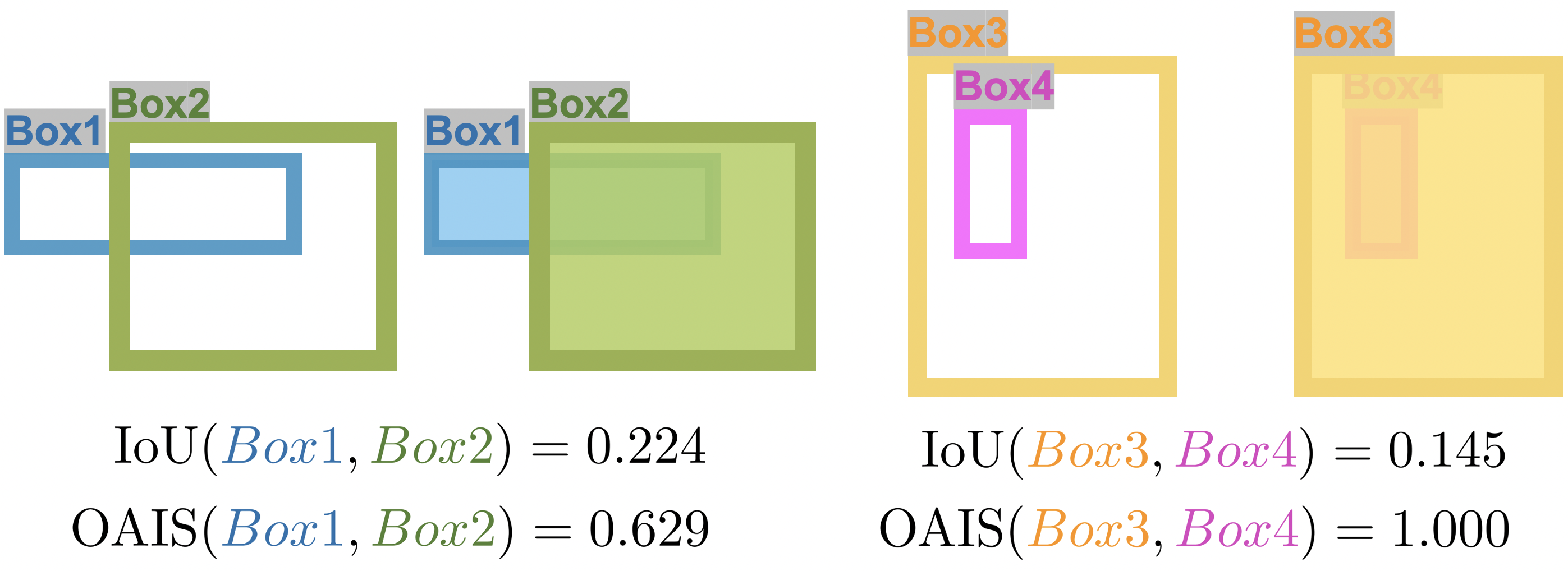}
    \caption{A schematic comparison of IoU and the proposed OAIS.} \label{fig:oais}  \vspace{-5pt}
\end{figure}

\subsection{Cross-Modal Data Augmentation}
\label{sec:cmaug}
\begin{table*}[hbt]
  \centering \small
  \tabcolsep=0.21cm
  \begin{tabular}{cccccccccccc}
    \toprule
    \multirow{2}{*}{Method} & \multirow{2}{*}{Venue} & \multirow{2}{*}{Extra Data} & \multicolumn{3}{c}{\textit{Test} $AP_{3D}$@IoU=0.7} & \multicolumn{3}{c}{\textit{Test} $AP_{BEV}$@IoU=0.7} & \multicolumn{3}{c}{\textit{Val} $AP_{3D}$@IoU=0.7} \\ 
    \cmidrule{4-6} \cmidrule{7-9} \cmidrule{10-12}
    & & & Easy & Mod. & Hard & Easy & Mod. & Hard & Easy & Mod. & Hard \\
    \midrule
    CaDDN \cite{caddn}  & CVPR'21  & LiDAR & 19.17 & 13.41 & 11.46 & 27.94 & 18.91 & 17.19 & 23.57 & 16.31 & 13.84 \\
    MonoFlex \cite{monoflex}& CVPR'21  & - & 19.94 & 13.89 & 12.07 & 28.23 & 19.75 & 16.89 & 23.64 & 17.51 & 14.83 \\
    GUPNet \cite{gupnet} & ICCV'21 & - & 20.11 & 14.20 & 11.77 & 30.29 & 21.19 & 18.20 & 22.76 & 16.46 & 13.72 \\
    MonoDTR \cite{monodtr} & CVPR'22 & LiDAR & 21.99 & 15.39 & 12.73 & 28.59 & 20.38 & 17.14 & 24.52 & 18.57 & 15.51 \\
    MonoDistill \cite{monodistill} & ICLR'22& LiDAR & 22.97 & 16.03 & 13.60 & 31.87 & 22.59 & 19.72 & 24.31 & 18.47 & 15.76 \\
    MonoJSG \cite{monojsg} & CVPR'22 & - & 24.69 & 16.14 & 13.64 & 32.59 & 21.26 & 18.18 & 26.40 & 18.30 & 15.40 \\
    DID-M3D \cite{didm3d} & ECCV'22 & - & 24.40 & 16.29 & 13.75 & 32.95 & 22.76 & 19.83 & 22.98 & 16.12 & 14.03 \\
    DD3D \cite{dd3d} & ICCV'21 & Depth & 23.22 & 16.34 & 14.20 & 30.98 & 22.56 & 20.03 & - & - & - \\
    MonoDETR \cite{monodetr} & ICCV'23 & - & 25.00 & 16.47 & 13.58 & 33.60 & 22.11 & 18.60 & 28.84 & 20.61 & 16.38\\
    ADD \cite{add} & AAAI'23 & LiDAR & 25.61 & 16.81 & 13.79 & 35.20 & 23.58 & 20.08 & 30.71 & 21.94 & 18.42\\
    MonoNeRD \cite{mononerd} & ICCV'23 & LiDAR & 22.75 & 17.13 & 15.63 & 31.13 & 23.46 & 20.97 & - & - & - \\
    MonoDDE \cite{monodde} & CVPR'22 & - & 24.93 & 17.14 & 15.10 & 33.58 & 23.46 & 20.37 & 26.66 & 19.75 & 16.72 \\
    MonoATT \cite{monoatt} & CVPR'23 & - & 24.72 & 17.37 & 15.00 & 36.87 & 24.42 & 21.88 & 29.56 & 22.47 & 18.65 \\
    DD3Dv2 \cite{dd3dv2} & ICRA'23  & LiDAR & 26.36 & 17.61 & 15.32 & 35.70 & 24.67 & 21.73 & - & - & - \\
    MoGDE \cite{mogde} & NeurIPS'22& - &27.07& 17.88 & 15.66 & 38.38 & 25.60 & \textbf{22.91} & - & - & - \\
    \midrule
    3DSeMo* \cite{3dsemo} & arXiv'23 & LiDAR & 23.55 & 15.25 & 13.24 & 30.99 & 21.78 & 18.64 & 27.35 & 20.87 & 17.66 \\
    LPCG* \cite{lpcg} & ECCV'22 & LiDAR & 25.56 & 17.80 & 15.38 & 35.96 & 24.81 & 21.86 & 31.15 & 23.42 & 20.60 \\
    Mix-Teaching* \cite{mixteaching} & CSVT'23 & LiDAR & 26.89 & 18.54 & 15.79 & 35.74 & 24.23 & 20.80 & 29.74 & 22.27 & 19.04 \\
    CMKD* \cite{cmkd} & ECCV'22 & LiDAR & 28.55 & 18.69 & 16.77 & 38.98 & 25.82 & 22.80 & 30.20 & 21.50 & 19.40 \\
    \midrule
    \textbf{ODM3D* (Ours)} & - & LiDAR & \textbf{29.75} & \textbf{19.09} &  \textbf{16.93} & \textbf{39.41} & \textbf{26.02} & 22.76 & \textbf{35.09} & \textbf{23.84} & \textbf{20.57} \\
    \rowcolor[HTML]{EDEDED}
    \textit{Improvements} & - & - & \textcolor[RGB]{61,145,64}{\textit{+1.20}} & \textcolor[RGB]{61,145,64}{\textit{+0.40}} & \textcolor[RGB]{61,145,64}{\textit{+0.16}} & \textcolor[RGB]{61,145,64}{\textit{+0.43}} & \textcolor[RGB]{61,145,64}{\textit{+0.20}} & \textcolor[RGB]{192,0,0}{\textit{-0.15}} & \textcolor[RGB]{61,145,64}{\textit{+4.89}} & \textcolor[RGB]{61,145,64}{\textit{+2.34}} & \textcolor[RGB]{61,145,64}{\textit{+1.17}} \\
    \bottomrule
  \end{tabular}
  \caption{$AP_{3D}|_{R_{40}}$ and $AP_{BEV}|_{R_{40}}$ results of ``Car" objects on KITTI \textit{test} and \textit{val} sets. * denotes semi-supervised methods. ``\textit{Improvements}" indicates absolute AP improvements compared to a CMKD baseline. Best results within each sub-category are marked in \textbf{bold}.}
  \label{tab:kittisemi} \vspace{-5pt}
\end{table*}

\noindent \textbf{IoU is flawed.} Existing cross-modal GT-sampling strategies for 3D object detection \cite{moca, autoalignv2, vff} universally adopt Intersection of Union (IoU) as a measure of the extent of overlap between pairs of objects in PV to avoid severe occlusion which harms training. However, we argue that IoU is a suboptimal criterion that often fails to indicate severe or even complete occlusion. In Fig. \ref{fig:cmaug}, an existing car in \textcolor[HTML]{3B71AA}{Box 1} is largely occluded by the pasted car in \textcolor[HTML]{5D803E}{Box 2}, leaving very limited visual cues for the monocular detector to learn from. The pasted pedestrian in \textcolor[HTML]{CB50BC}{Box 4} is almost entirely occluded by the car in \textcolor[HTML]{5D803E}{Box 2} and indeed entirely occluded by another pasted pedestrian in \textcolor[HTML]{F09837}{Box 3}. These severe occlusion cases successfully passed collision tests with an IoU threshold of 0.5. They cause objects having very limited or zero pixels to remain in the augmented scene, and only serve to mislead and harm the monocular detector's learning. 

\noindent \textbf{An occlusion-aware criterion.} The root of IoU's malfunction lies in its inability to reason about relative depth of boxes. In simple words, IoU measures the extent of overlap of boxes on a 2D plane, but is unaware of which box is being occluded by which in 3D space. We avoid this shortcoming of IoU by introducing a novel Occlusion-Aware Intersection Score (OAIS), which instead calculates the intersection \textit{over the area of the 2D box that has a larger depth value}. Mathematically:
\begin{equation} \small
    \mathrm{OAIS}(B1, B2) = \frac{\mathrm{Area}(B1 \cap B2)}{\mathrm{Area}(\mathrm{Max_{D}}(B1, B2)))}
\end{equation}
where $B1$ and $B2$ are 2D boxes projected from 3D boxes with respective depth values; $\mathrm{Max_{D}(\cdot)}$ is an operator that selects the box with a larger depth value (a random selection if equal depth values). For the case $B1 \cap B2 \neq \varnothing $, this translates into intersection \textit{over area of the box being occluded}. In practice, we take the depth of the ground-truth 3D bounding box from which a 2D box is projected as the 2D box's depth.

As shown in Fig.~\ref{fig:oais}, when using IoU as the collision metric, significantly occluded \textcolor[HTML]{3B71AA}{Box 1} has a low IoU of $0.224$ with \textcolor[HTML]{5D803E}{Box 2}. \textcolor[HTML]{CB50BC}{Box 4} has an even lower IoU of $0.145$ with \textcolor[HTML]{F09837}{Box 3} despite being fully occluded by the latter. In comparison, OAIS between \textcolor[HTML]{3B71AA}{Box 1} and \textcolor[HTML]{5D803E}{Box 2} is $0.629$ which provides an intuitive measure of how much of \textcolor[HTML]{3B71AA}{Box 1} has been occluded.  \textcolor[HTML]{F09837}{Box 3} and \textcolor[HTML]{CB50BC}{Box 4} yield a maximal OAIS of $1.0$, implying that ``whichever box being occluded has itself 100\% occluded". Under a nominal collision threshold of $0.5$, these two severe occlusion cases are kept when using IoU, but would have been avoided with the proposed OAIS.

\begin{figure}
    \centering \captionsetup{skip=5pt}
    \includegraphics[width=\columnwidth]{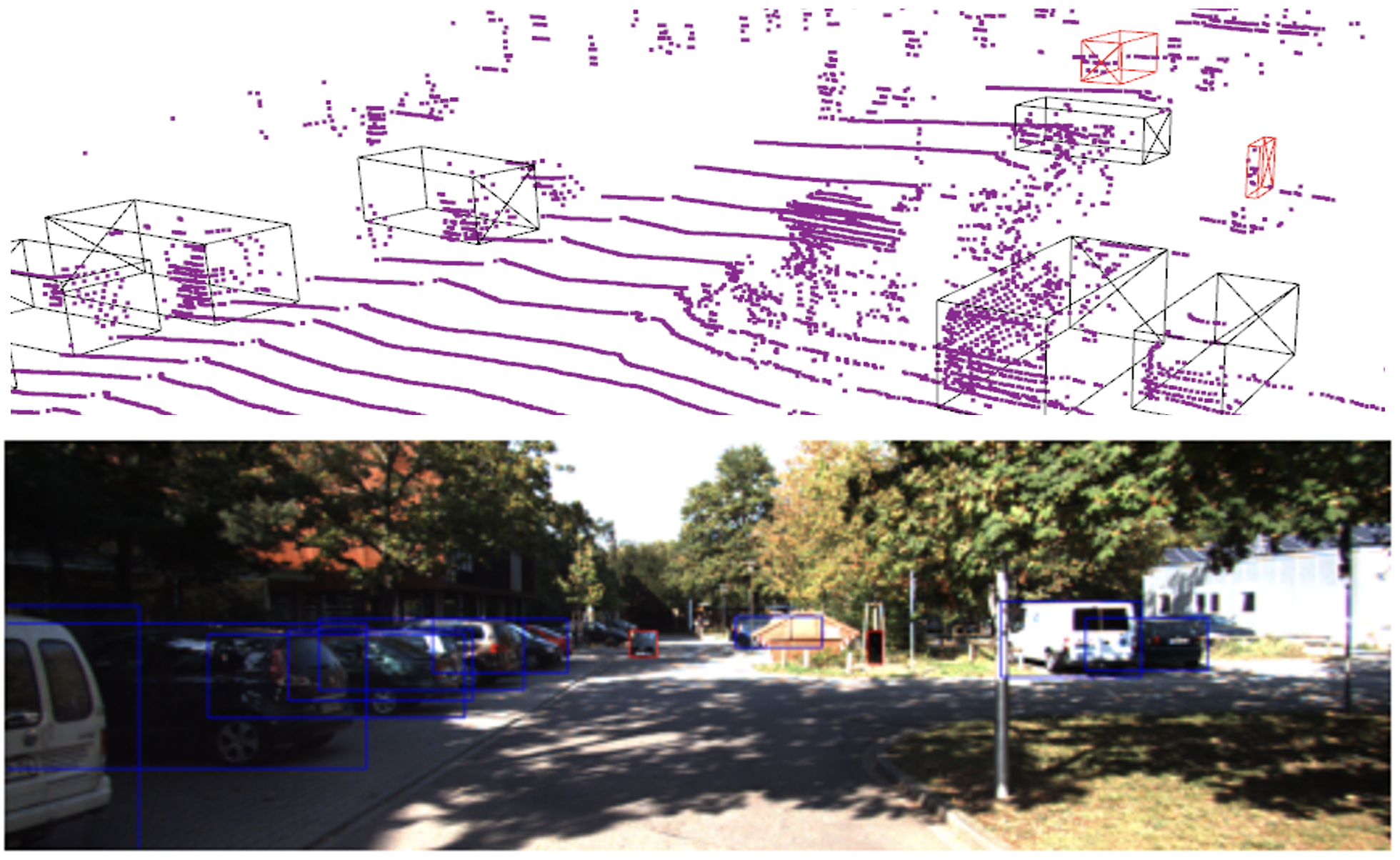}
    \caption{A pasted object (marked with red boxes) can result in an adequately distinguishable cluster of points in LiDAR but a tiny patch in the image. }
    \label{fig:pv_size} \vspace{-10pt}
\end{figure}

\noindent \textbf{Filtering objects by PV size. }
We further observed experimentally that excessively tiny patches pasted into images can harm the training of the monocular detector. Traditionally, GT-sampling \cite{second} filters off objects with very few LiDAR points (\eg $< 5$ points). Yet, we observed that while a faraway object may very well contain a dozen of points, clearly distinguishable in the LiDAR scene, it can occupy a rather limited number of pixels in the image (\eg pasted ``Pedestrian" in Fig. \ref{fig:pv_size} takes up $30\times13$ pixels - $0.083\%$ of the entire PV space) due to perspective, which is further exacerbated when occluded by other pasted objects. Therefore, we design an extra filter to prevent objects excessively small in PV from being pasted into the image. As shown in Sec.~\ref{sec:ablation_cmaug}, this simple design leads to a further $0.3 AP_{3D}$ increase on moderate cars.

\noindent \textbf{Pseudo-labels for collision tests.} It is noteworthy that collision tests demand knowledge on the location of objects existing in a scene. This has been conveniently obtained from the ground-truth annotations of labelled data in prior works \cite{moca, autoalignv2, vff, focalsconv, logonet}. Our semi-supervised setting, however, involves large amounts of unlabelled scenes. To acquire the rough location of unannotated objects, we apply the pre-trained teacher for inference on unlabelled scenes and utilise the generated pseudo-labels for collision tests in CMAug. During the teacher's inference, we adopt a lower confidence threshold to discourage false negative detections, since a missed detection may cause intermingled existing and pasted objects in both images and point clouds.

\noindent \textbf{CMAug workflow.} Finally, we outline the procedures of our proposed CMAug strategy. Prior to cross-modal distillation, we first build a database of all objects in labelled training samples, similar to \cite{second}. Next, we generate pseudo-labels for all unlabelled training samples and store them as their labels. Afterwards, cross-modal distillation training starts and we randomly select an arbitrary number of objects from the object database to paste into each training scene. Objects whose projected box in PV is less than a pre-defined size are discarded. Remaining object points are pasted into the scene's point cloud, and object patches into the image, using 3D and projected 2D bounding boxes, respectively. Each time a new group of objects is sampled, point cloud collision tests are conducted in BEV using IoU, and image patch collision tests in PV using OAIS, between each pair of existing objects and to-be-pasted objects as well as among all to-be-pasted objects. Objects that fail any collision tests will be discarded. Eventually, all kept objects are pasted into the scene in a far-to-near order. In Sec.~\ref{sec:ablation}, we show that CMAug also generalises beyond cross-modal distillation and M3OD.

\begin{figure}[t]
    \centering 
    \begin{subfigure}{\columnwidth}
        \centering \captionsetup{skip=0pt}
        \includegraphics[width=\columnwidth]{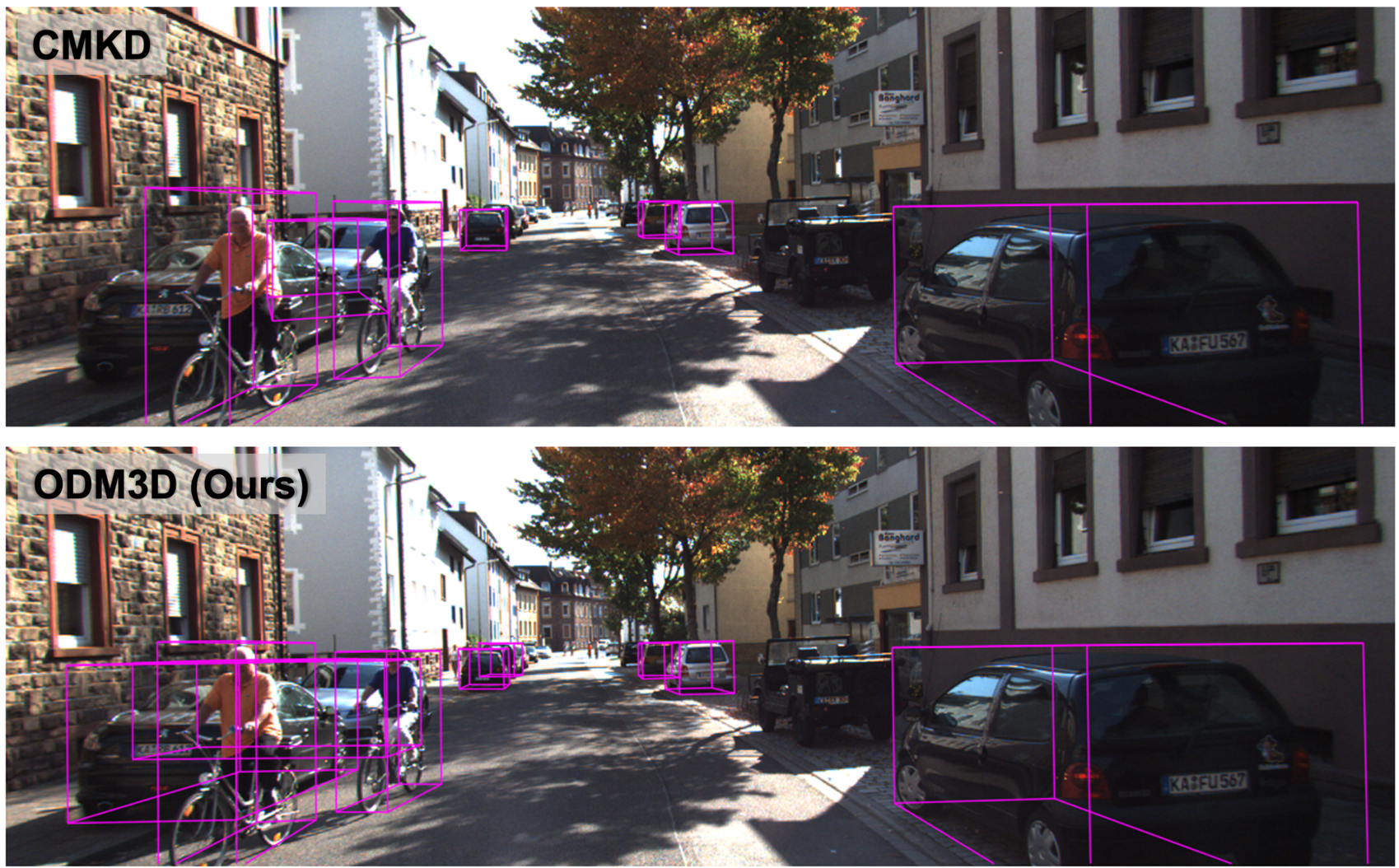}
        \end{subfigure}
        \begin{subfigure}{\columnwidth}
        \includegraphics[width=\columnwidth]{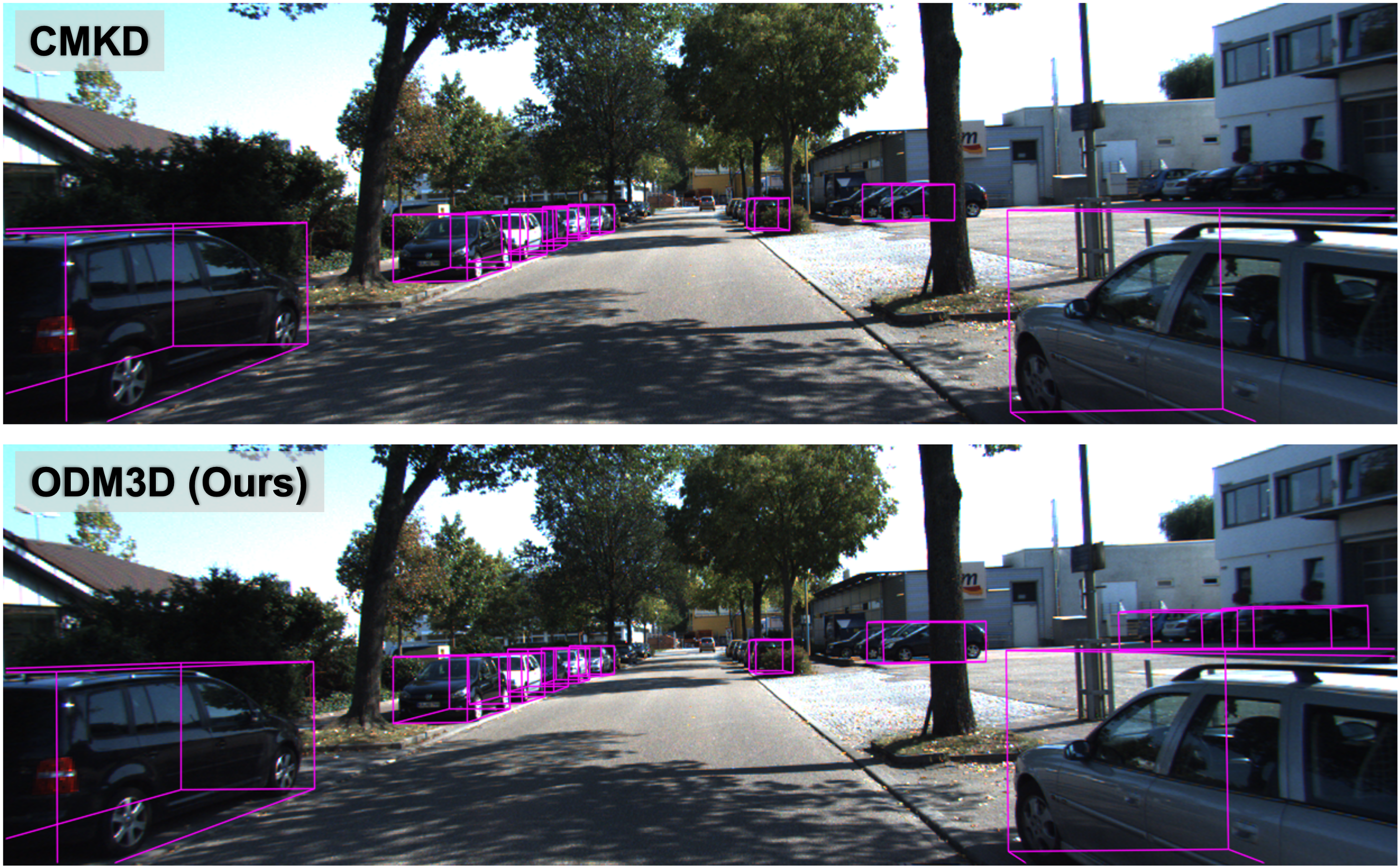}
        \end{subfigure}
    \caption{Qualitative comparison of detection results by our method and CMKD \cite{cmkd}.} \label{fig:qualitative_results}
    \vspace{-10pt}
\end{figure}

\section{Experiments}
\subsection{Implementation Details}
\noindent \textbf{Datasets.} 
We validate our framework on the KITTI 3D \cite{kitti3d} dataset, which consists of $7{\small,}481$ training and $7{\small,}518$ test images with corresponding point clouds. For local evaluation, we follow the convention to divide training images into a training subset of $3{\small,}712$ images and a validation set of $3{\small,}769$ images, dubbed KITTI \textit{train} and \textit{val}, respectively. The best model determined by KITTI \textit{val} is evaluated on the test set, denoted as KITTI \textit{test}. 
Objects in KITTI 3D are annotated into three difficulty levels: ``Easy", ``Moderate", and ``Hard", with Average Precision (AP) as the official evaluation metric. KITTI 3D is an annotated subset of the KITTI Raw dataset \cite{kittiraw}, which further comprises around 42k unannotated images and corresponding point clouds in sequence form, which are exploited under our semi-supervised learning framework. For validation, we follow \cite{geomconsist, cmkd} and use the \texttt{eigen-clean} subset \cite{simonelli21} of KITTI Raw to avoid data leakage due to scenes overlapping with KITTI \textit{val}.

\noindent \textbf{Network details.} 
We choose the state-of-the-art CMKD \cite{cmkd} as our baseline and implement our framework based on the OpenPCDet \cite{openpcdet} codebase. We use SECOND \cite{second} as our LiDAR-based teacher and CaDDN \cite{caddn} our monocular student. 
Our CaDDN student follows the same settings as in \cite{caddn} and \cite{cmkd}, except that depth maps are not utilised for supervising categorical depth estimation since our training scenes have been altered by CMAug. Instead, depth estimation is supervised implicitly by dense distillation.

\noindent \textbf{Training details.} 
Our framework is trained on a single NVIDIA RTX 3090 GPU with a batch size of 4. We follow a two-stage distillation strategy and train the framework for 30 epochs in stage 1 ($\mathcal{L_{\mathrm{FeatKD}}}$ only) and 15 epochs in stage 2 ($\mathcal{L_{\mathrm{FeatKD}}}$ and $\mathcal{L_{\mathrm{RespKD}}}$), using both labelled and unlabelled data. The proposed CMAug is applied in both stages. More experimental details are provided in the supplementary material.

\subsection{Comparisons with Prior Arts}
\noindent \textbf{Quantitative results.} We make a detailed quantitative comparison of our method against recently published supervised and semi-supervised monocular 3D object detectors on both KITTI \textit{test} and \textit{val} sets. We report the results on the ``Car" category since it is the most important category in the KITTI 3D dataset (``Pedestrian" and ``Cyclist" results are provided in the supplementary material). As shown in Tab. \ref{tab:kittisemi}, our method drastically boosts a CaDDN \cite{caddn} model from $AP_{3D}\,  13.41$ to $19.09$ (a 42.4\% increase) on moderate cars owing to effective usage of extra LiDAR and unlabelled data. Our method surpasses all existing methods, supervised or semi-supervised, by considerable margins, including the current state-of-the-art CMKD \cite{cmkd}. Specifically, on KITTI \textit{test}, ODM3D outperforms CMKD by $0.40$ and $0.20$ on $AP_{3D}$ and $AP_{BEV}$, respectively. Larger performance gains are observed on KITTI \textit{val}, where ODM3D is ahead of CMKD by $2.34\, AP_{3D}$ on moderate cars and outperforms all other methods by at least $0.4\, AP_{3D}$.

\noindent \textbf{Qualitative results.} Fig. \ref{fig:qualitative_results} visualises detections by our method and CMKD. In the first scene, it is clear that ODM3D precisely detects the two cars on the left and another car further down the street which are missed by CMKD. In the second scene, ODM3D picks up two cars on the right and one occluded by a tree with higher accuracy. These examples show that our method better handles faraway and occluded objects, which is likely owing to our occlusion-aware augmentation and foreground-attentive occupancy-guided distillation.

\begin{table}[t]
  \centering \small \captionsetup{skip=6pt}
  \tabcolsep=0.12cm
  \begin{tabular}{cccccccccc}
  \toprule
   \multirow{2}{*}{Expt.} & \multirow{2}{*}{FD} & \multirow{2}{*}{RD} & \multirow{2}{*}{O-FD} & \multirow{2}{*}{O-RD} & \multirow{2}{*}{CMA} & \multicolumn{3}{c}{\textit{Val} $AP_{3D}$@IoU=0.7}\\ \cmidrule{7-9}
 & &  &  &  & & Easy & Mod. & Hard \\ \hline
1 & \checkmark  & \checkmark &  & &  & 32.67 & 21.54 & 18.79 \\
2 & \checkmark  & \checkmark &  & & \checkmark & 31.50 & 22.18 & 19.34 \\
3 & \checkmark &  & & \checkmark & & 33.02 & 21.89 & 18.19 \\
4 & & \checkmark & \checkmark &  & & 34.69 & 23.68 & 20.55 & \\
5 & & & \checkmark & \checkmark & & 34.84 & 23.77 & 20.04 \\
6 & & & \checkmark & \checkmark & \checkmark & \textbf{35.09} & \textbf{23.84} & \textbf{20.57} \\
   \bottomrule
   \end{tabular}
\caption{Ablation experiments on core components of our method. ``FD" and ``RD" stand for vanilla feature and response distillation, respectively. ``O-" denotes their occupancy-guided variants. ``CMA" denotes the proposed CMAug.}
\label{tab:ablation_main} \vspace{-7pt}
\end{table}

\begin{table}[t]
  \centering \small \captionsetup{skip=6pt}
  \tabcolsep=0.09cm
  \begin{tabular}{cccccccc}
  \toprule
   \multirow{2}{*}{Expt.} & \multirow{2}{*}{OFD-V} & \multirow{2}{*}{ORD-V} & \multirow{2}{*}{OFD-G} & \multirow{2}{*}{ORD-G} & \multicolumn{3}{c}{\textit{Val} $AP_{3D}$@IoU=0.7}\\ \cmidrule{6-8}
 & & &  & & Easy & Mod. & Hard \\ \hline
1 & \checkmark & \checkmark & & & 34.11 & 23.36 & 19.50 \\
2 & \checkmark & & & \checkmark & 34.58 & 23.72 & 19.83 \\
3 & & \checkmark & \checkmark & & 34.63 & 23.59 & 19.85\\
4 & & & \checkmark & \checkmark & \textbf{34.84} & \textbf{23.77} & \textbf{20.04} \\
   \bottomrule
   \end{tabular}
\caption{Ablation experiments on the use of Gaussian smoothing in occupancy-guided feature and response distillation. ``OFD-V" and ``ORD-V" denote vanilla occupancy-guided feature and response distillation, respectively, whereas ``-G" indicates their variants using Gaussian-smoothed occupancy masks.}
\label{tab:ablation_gaussian} \vspace{-7pt}
\end{table}

\begin{table}[t]
  \centering \small \captionsetup{skip=6pt}
  \tabcolsep=0.08cm
  \begin{tabular}{cccccccccc}
  \toprule
   \multirow{2}{*}{Expt.} & \multirow{2}{*}{CD} & \multirow{2}{*}{LD} & \multirow{2}{*}{DD} & \multirow{2}{*}{CD-G} & \multirow{2}{*}{LD-G} & \multirow{2}{*}{DD-G} & \multicolumn{3}{c}{\textit{Val} $AP_{3D}$@IoU=0.7}\\ \cmidrule{8-10}
 & & & & & & & Easy & Mod. & Hard \\ \hline
1 & \checkmark & \checkmark & \checkmark & & & & 34.63 & 23.59 & 19.85 \\
2 & & \checkmark & \checkmark & \checkmark & & & 34.66 & 23.71 & 19.98\\
3 &  \checkmark & & \checkmark & & \checkmark & & 34.50 & 23.75 & 19.81\\
4 & & & \checkmark & \checkmark & \checkmark & & \textbf{34.97} & 23.75 & 20.03 \\
5 & & & & \checkmark & \checkmark & \checkmark & 34.84 & \textbf{23.77} & \textbf{20.04} \\
   \bottomrule
   \end{tabular}
\caption{Ablation experiments on the use of Gaussian smoothing for various heads in occupancy-guided response distillation. ``CD", ``LD" and ``DD" denote vanilla classification, localisation and direction distillation, respectively, whereas ``-G" denotes their variants using Gaussian-smoothed occupancy masks.}
\label{tab:ablation_rkd} \vspace{-7pt}
\end{table}

\subsection{Ablation Studies}
\label{sec:ablation}
\noindent \textbf{Effectiveness of core designs.} 
Tab. \ref{tab:ablation_main} studies the effectiveness of each core component of our framework. It is clear that the proposed occupancy-guided feature distillation and occupancy-guided response distillation lead to increased detection results both individually (Expt. 1$\rightarrow$3, 4) and collectively (Expt. 1$\rightarrow$5) compared to  baseline results (Expt. 1). It can be inferred from Expt. 3 and 4 that improvements brought about by the use of BEV occupancy guidance (Expt. 1$\rightarrow$5) are primarily accounted for by occupancy-guided feature distillation. This also suggests that high-quality BEV feature maps can be a prerequisite for accurate and effective response distillation and detection that take place downstream. The proposed CMAug leads to improved results on both CMKD \cite{cmkd} (Expt. 1$\rightarrow$2) and occupancy-guided distillation (Expt. 5$\rightarrow$6) baselines, highlighting its effectiveness and potential as a versatile, plug-and-play gadget for boosting joint LiDAR-RGB learning.  
Finally, the highest detection results are achieved using occupancy-guided feature distillation, occupancy-guided response distillation, and CMAug altogether (Expt. 6).

\noindent \textbf{Effectiveness of occupancy-guided distillation designs.}
\label{sec:ablation_okd} 
The benefits of smoothed BEV occupancy masks are illustrated in ablation experiments in Tab. \ref{tab:ablation_gaussian}. Performance drops take place with Gaussian smoothing ablated in either feature (Expt. 4$\rightarrow$2) or response (Expt. 4$\rightarrow$3) distillation, and worst performance is observed when it is absent in both distillations (Expt. 4$\rightarrow$1). We further study the effect of Gaussian smoothing in distilling each detection head in Tab. \ref{tab:ablation_rkd}. The results show that smoothed occupancy masks result in improved detection when applied either individually to each detection head (Expt. 1$\rightarrow$2,3) or to multiple heads in combinatorial ways (Expt. 1,2,3$\rightarrow$4,5). 

\noindent \textbf{Effectiveness of CMAug designs.} 
\label{sec:ablation_cmaug}
From Tab. \ref{tab:ablation_cmaug}, naively applying the commonly adopted MixedAug \cite{vff}, which performs simple IoU-based collision tests, leads to significantly degraded detection performance. Replacing the IoU score with the proposed OAIS immediately gives rise to drastic increases in performance (a notable 21.2\% increase on moderate cars), turning data augmentation's contribution from negative to positive and surpassing the baseline. Besides, filtering by PV sizes also boosts detection performance on moderate and hard cars.

\noindent \textbf{Generalisation studies of CMAug.}
We further apply our CMAug to representative and high-performance multi-modal 3D object detectors VFF \cite{vff}, FocalsConv \cite{focalsconv} and LoGoNet
\cite{logonet}. As shown in Tab. \ref{tab:aug_generalisation}, by simply replacing their default augmentation, the proposed CMAug consistently boosts the detection accuracy of these multi-modal detectors, with all other settings unchanged. These results corroborate our previous arguments that the identified ``IoU defect" is universal among 3D object detectors employing cross-modality GT-sampling, and our proposed fix to it generalises beyond monocular detectors.
\vspace{-5pt}

\begin{table}[t]
  \centering \small \captionsetup{skip=6pt}
  \begin{tabular}{cccc}
    \toprule
    \multirow{2}{*}{Method} & \multicolumn{3}{c}{\textit{Val} $AP_{3D}$@IoU=0.7} \\ 
    \cmidrule{2-4} \
    & Easy & Mod. & Hard   \\
    \midrule
    Baseline & \textbf{32.67} & 21.54 & 18.79 \\
    + MixedAug \cite{vff}& 27.77 & 18.58 & 16.44 \\
    \rowcolor[HTML]{EDEDED}
    \textit{Improvements} &  \textcolor[RGB]{192,0,0}{\textit{-4.90}} & \textcolor[RGB]{192,0,0}{\textit{-2.96}} &  \textcolor[RGB]{192,0,0}{\textit{-2.35}} \\    
    \midrule
    + OAIS & 32.42 & 21.87 & 18.99 \\
    + MinPxFilter & 32.57 & \textbf{22.21} & \textbf{19.43} \\
    \rowcolor[HTML]{EDEDED}
    \textit{Improvements} & \textcolor[RGB]{192,0,0}{\textit{-0.10}} & \textcolor[RGB]{61,145,64}{\textit{+0.76}} & \textcolor[RGB]{61,145,64}{\textit{+0.64}} \\    
    \bottomrule
  \end{tabular}
  \caption{Ablation experiments on the components of CMAug and performance comparison between CMAug and MixedAug \cite{vff}.}
  \label{tab:ablation_cmaug} \vspace{-5pt}
\end{table}

\begin{table}[t]
  \centering \small \captionsetup{skip=6pt}
  \begin{tabular}{cccc}
    \toprule
    \multirow{2}{*}{Method} & \multicolumn{3}{c}{\textit{Val} $AP_{3D}$@IoU=0.7} \\ 
    \cmidrule{2-4} \
    & Easy & Mod. & Hard   \\
    \midrule
    VFF--Voxel-RCNN \cite{vff} & \textbf{92.80} & 83.53 & 82.78 \\
    + CMAug  & 92.64 & \textbf{83.68} & \textbf{82.97} \\
    \rowcolor[HTML]{EDEDED}
    \textit{Improvements} & \textcolor[RGB]{192,0,0}{\textit{-0.16}} & \textcolor[RGB]{61,145,64}{\textit{+0.15}} & \textcolor[RGB]{61,145,64}{\textit{+0.19}} \\  
    \midrule
    FocalsConv--PV-RCNN \cite{focalsconv} & 91.89 & 85.31 & 83.10 \\
    + CMAug & \textbf{92.55} & \textbf{85.53} & \textbf{83.33} \\
    \rowcolor[HTML]{EDEDED}
    \textit{Improvements} & \textcolor[RGB]{61,145,64}{\textit{+0.66}} & \textcolor[RGB]{61,145,64}{\textit{+0.22}} & \textcolor[RGB]{61,145,64}{\textit{+0.23}} \\
    \midrule
    LoGoNet \cite{logonet} & 91.92 & 85.02 & 82.88 \\
    + CMAug & \textbf{92.16} & \textbf{85.20} & \textbf{83.02} \\
    \rowcolor[HTML]{EDEDED}
    \textit{Improvements} & \textcolor[RGB]{61,145,64}{\textit{+0.24}} & \textcolor[RGB]{61,145,64}{\textit{+0.18}} & \textcolor[RGB]{61,145,64}{\textit{+0.14}} \\    
    \bottomrule
  \end{tabular}
  \caption{A generalisation study on CMAug applied to multi-modal 3D object detectors. Baseline results are produced from official code for a fair assessment of CMAug's effectiveness.} 
  \label{tab:aug_generalisation} \vspace{-5pt}
\end{table}

\section{Conclusion}
\label{sec::conclusion}
In this paper, we proposed ODM3D, a novel knowledge distillation framework that alleviates the foreground sparsity issue in autonomous driving scenes for enhanced semi-supervised monocular 3D object detection. We showed that exploiting the inherent ground-truth 3D occupancy knowledge in point clouds significantly benefits knowledge distillation in both feature and prediction spaces, as the network is encouraged to attend to regions that more likely contain objects. We also demonstrated that our proposed cross-modal data augmentation strategy not only enriches supervisory signals throughout the cross-modality learning process, but also generates more realistic and learner-friendly augmented scenes. Extensive experiments on the KITTI dataset have validated the effectiveness of the proposed method.

{\small
\bibliographystyle{ieee_fullname}
\bibliography{egbib}
}

\end{document}